\documentclass[10pt,twocolumn,letterpaper]{article}

\usepackage{iccv}              

\usepackage{url}
\usepackage{graphicx}
\usepackage{wrapfig}
\usepackage{booktabs}
\usepackage{float}
\usepackage{caption}
\usepackage{makecell}
\usepackage{multirow}
\usepackage{cuted}
\usepackage{bm}

\definecolor{iccvblue}{rgb}{0.21,0.49,0.74}
\usepackage[pagebackref,breaklinks,colorlinks,allcolors=iccvblue]{hyperref}

\def\paperID{****} 
\def\confName{ICCV}
\def\confYear{2025}

\title{LucidFusion: Reconstructing 3D Gaussians with Arbitrary Unposed Images}

\author{
    \parbox{\textwidth}{\centering
        Hao He\textsuperscript{1,2}\thanks{Equal contribution.} \quad
        Yixun Liang\textsuperscript{2}\footnotemark[1] \quad
        Luozhou Wang\textsuperscript{1} \quad
        Yuanhao Cai\textsuperscript{3} \quad
        Xinli Xu\textsuperscript{1} \\
        Hao-Xiang Guo\textsuperscript{4} \quad
        Xiang Wen\textsuperscript{4} \quad
        Yingcong Chen\textsuperscript{1,2}\thanks{Corresponding author.} \\
        \tt\small$^1$ HKUST(GZ) \quad
        \tt\small$^2$ HKUST \quad
        \tt\small$^3$ Johns Hopkins University \quad
        \tt\small$^4$ SkyWork AI \\
    }
}

\begin{document}
\maketitle

\begin{strip}
    \centering
    \vspace{-4em}
    \includegraphics[width=\textwidth]{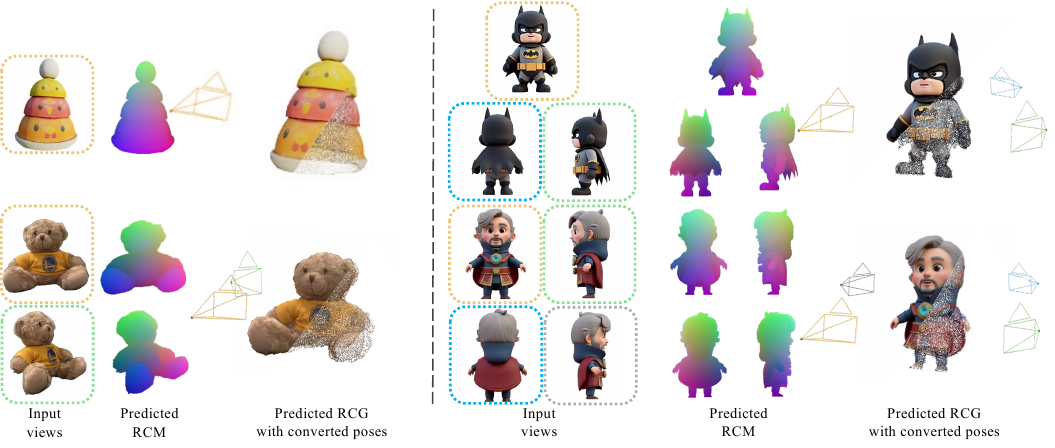}
    \captionsetup{font=small}
    \captionof{figure}{LucidFusion utilizes \textit{Relative Coordinate Gaussian (RCG)} representation to achieve 3D reconstruction with pose estimation from unposed, sparse and arbitrary numbers of input views in a feed-forward manner. }
    \label{fig:teaser}
\end{strip}
\vspace{-4em}

\begin{abstract}
Recent large reconstruction models have made notable progress in generating high-quality 3D objects from single images. However, current reconstruction methods often rely on explicit camera pose estimation or fixed viewpoints, restricting their flexibility and practical applicability. We reformulate 3D reconstruction as image-to-image translation and introduce the \textit{Relative Coordinate Map (RCM)}, which aligns multiple unposed images to a “main” view without pose estimation. While RCM simplifies the process, its lack of global 3D supervision can yield noisy outputs. To address this, we propose \textit{Relative Coordinate Gaussians (RCG)} as an extension to RCM, which treats each pixel’s coordinates as a Gaussian center and employs differentiable rasterization for consistent geometry and pose recovery. Our LucidFusion framework handles an arbitrary number of unposed inputs, producing robust 3D reconstructions within seconds and paving the way for more flexible, pose-free 3D pipelines.
\end{abstract}

\section{Introduction}\label{sec:intro}

Digital 3D objects are increasingly essential in a variety of domains, facilitating immersive visualization, analysis, and interaction with objects and environments that closely mimic real-world experiences. These objects are foundational in fields such as architecture, animation, gaming, and virtual and augmented reality, with broad applications across industries like retail, online conferencing, and education. Despite their growing demand, producing high-quality 3D content remains a resource-intensive task, requiring substantial time, effort, and domain expertise. This challenge has catalyzed the rapid advancement of 3D content generation techniques~\cite{mildenhall2021nerf,wang2021nerf,he2023cp,hong2023lrm,zou2024triplane,huang2024zeroshape,liang2024luciddreamer,wang2024prolificdreamer}, including methods that reconstruct 3D objects from one or more input images.

Recently, 3D reconstruction methods~\cite{hong2023lrm, tochilkin2024triposr, zou2024triplane} have gained considerable attention, as they can convert single or multiple images, either captured by external device or generated by diffusion models, into complete 3D objects in content generation workflows. 
However, these methods inevitably require camera pose as an intermediate step to map image features into 3D: whether explicitly, as in traditional MVS-based methods~\cite{yao2018mvsnet, chen2021mvsnerf}, or implicit, as in LRM-based approaches~\cite{hong2023lrm}.
Yet, obtaining accurate poses of the input views is a non-trivial task: current methods often rely on external pose estimation pipelines (e.g., COLMAP~\cite{schonberger2016structure}) or fix the input viewpoint~\cite{tang2024lgm}, substantially constraining both the flexibility of the reconstruction process and user experience. 

\begin{figure}[!t]  
    \centering
    \includegraphics[width=\linewidth]{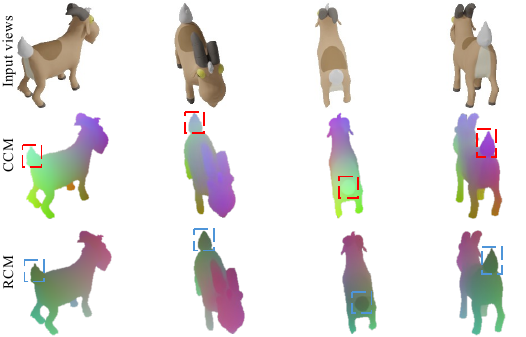}
    \captionsetup{font=small}
    \caption{Pilot study. We compare CCM and RCM given a set of input images. CCM fails to maintain consistency across different input views, as shown in the red box, while RCM successfully maintains the 2D-3D relation, as shown in blue box.}
    \label{fig:ccm_vs_rcm}
\end{figure}

This observation raises a critical problem: \textit{Can we mitigate the pose requirement for 3D reconstruction?} By revisiting the reconstruction problem (detailed in Sec.~\ref{sec:preliminary}), we find that the wrapping from 2D to 3D can be learned via an image-to-image translation approach, if we leverage an intermediate representation such as Canonical Coordinate Maps (CCM)~\cite{li2023sweetdreamer, wang2024crm}, we can bypass common challenges associated with pose estimation, allowing a more flexible 3D reconstruction pipeline. 
However, in practice, CCMs are difficult to regress because orientation cues are only implicitly embedded in the color space and such "orientation" information is not well-defined, as shown in Fig.~\ref{fig:ccm_vs_rcm}. 
To address this shortcoming, we propose the \textit{Relative Coordinate Map (RCM)}, which transforms each pixel to the camera space of a selected "main" camera (e.g., the first frame in our system), as shown in Fig.~\ref{fig:teaser}. 
This simple yet effective modification retrains CCM's advantage of end-to-end learnability via an image-to-image framework, while mitigating the ambiguities that arise from implicitly encoded orientation information.

Nevertheless, we observe that naively performing this mapping often results in inconsistent and noisy outputs as shown in Fig.~\ref{fig:stage_1_2_pc_compare}, primarily due to the lack of 3D prior supervision. To address this limitation, we further introduce \textit{Relative Coordinate Gaussians (RCG)}, interpreting each pixel's coordinates as the center of a Gaussians. 
The RCG extension allows differentiable rasterization from arbitrary viewpoints, and can be supervised by ground-truth images rather than solely by per-view coordinate predictions. This additional supervision resolves the noise and misalignment issues that arise under purely RCM training. 
By re-framing the multi-view reconstruction problem as an image-to-RCG transformation, we can efficiently obtain complete 3D representation from arbitrary, unposed images, as shown in Fig.~\ref{fig:teaser}. Furthermore, since RCG is inherently a 3D representation, it eliminates the common challenge associated with pose estimation and can directly recover camera poses. This feature is often missing from other feedforward, Gaussian-based methods.
In summary, our contributions are threefold:
\begin{itemize}
\item We revisit the reconstruction problem and identify the gap in existing CCM approaches, leading to the proposed RCM and its RCG extension.
\item We develop a system, \textit{LucidFusion}, that efficiently maps images to RCG, embedding pixel-wise correspondences across different views into a “main” view and eliminating explicit pose estimation.
\item We showcase the superior quality and flexibility of our method, enabling rapid 3D reconstruction and pose estimation within seconds.
\end{itemize}

\begin{figure*}[!t]
    \centering
    \includegraphics[width=\textwidth]{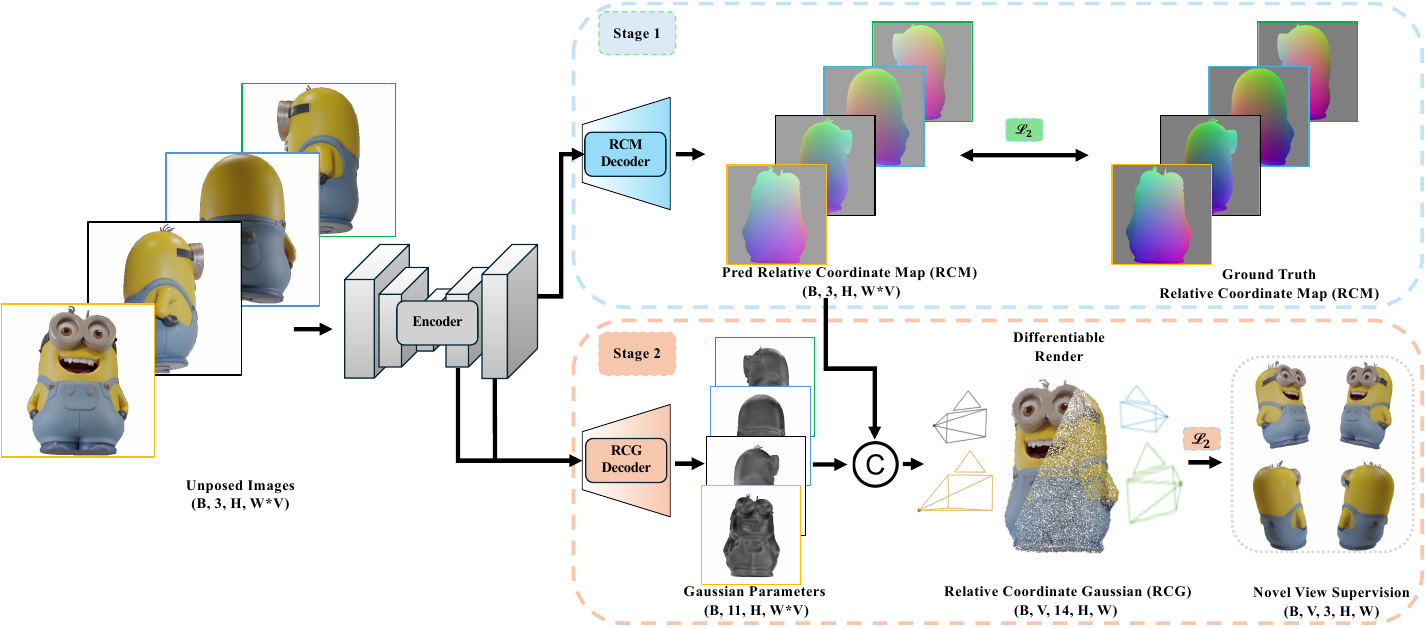}
    \captionsetup{font=small}
    \caption{Pipeline Overview of LucidFusion. Our framework processes a set of sparse, unposed multi-view images as input. The model predicts the RCM representation for the input images. Additionally, the feature map from the final layer of the encoder is fed into a decoder network to extend the RCM representation to RCG. The RCG is then rendered at novel views and supervised with ground truth images.}
    \label{fig:model_pipeline}
\end{figure*}

\section{Related Work}\label{sec:related}
\subsection{Multi-View 3D Reconstruction} 
Multi-view 3D reconstruction typically relies on multi-view stereo (MVS), which reconstructs the visible surface of an object by triangulating between multiple views. MVS-based methods can be broadly classified into three categories: depth map-based methods~\cite{campbell2008using, schonberger2016pixelwise, chang2022rc, ren2023volrecon, liang2024retr}, voxel grid-based methods~\cite{kutulakos2000theory, yao2019recurrent, chen2021mvsnerf}, and point cloud-based methods~\cite{furukawa2009accurate, chen2019point}. These methods generally operate by taking multi-view images and constructing a 3D cost volume through the unprojection of 2D multi-view features into plane sweeps. However, they all depend on the availability of camera parameters with the input multi-view images, either provided during data acquisition or estimated using Structure-from-Motion (SFM)~\cite{schonberger2016structure, jiang2013global} for in-the-wild reconstructions. Consequently, these methods often fail when handling sparse-view inputs without known camera poses. In contrast, our approach leverages the RCM representation, enabling 3D generation from uncalibrated and unposed sparse inputs, thereby offering a robust solution for real-world applications.

\subsection{Radiance Field Reconstruction}
Neural radiance fields (NeRF)~\cite{mildenhall2021nerf} have recently driven significant advancements in radiance field methods, achieving state-of-the-art performance~\cite{chen2021mvsnerf, wang2021ibrnet, ge2023ref}. These approaches optimize radiance field representations through differentiable rendering, diverging from traditional MVS pipelines, yet they still rely on dense sampling for precise reconstruction. To address sparse-view challenges in NeRF, recent works have incorporated regularization terms~\cite{niemeyer2022regnerf, wang2023sparsenerf} or leveraged geometric priors~\cite{chen2021mvsnerf, yang2023freenerf}. However, these methods continue to require image samples with known camera poses. Another research direction explores SDS-based optimization techniques, distilling detailed information from 2D diffusion models into 3D representations~\cite{poole2022dreamfusion, wang2024prolificdreamer, liang2024luciddreamer}, which enables the rendering of high-fidelity scenes but requires lengthy optimization for each individual scene. In contrast, our approach eliminates the need for known camera poses and operates in a feed-forward manner, supporting generalizable 3D generation without extensive optimization.

\subsection{Unconstrained Reconstruction}
Large Reconstruction Model (LRM)~\cite{hong2023lrm} introduced a triplane-based approach combined with volume rendering, demonstrating that a regression model can robustly predict a neural radiance field from a single-view image and thus reduce dependence on camera poses. Subsequent works~\cite{li2023instant,shi2023zero123++, shi2023mvdream, xu2023dmv3d, tang2024lgm, zhang2024gs} have leveraged diffusion models to extend single-view inputs to multi-view inputs, bypassing the need for camera poses. However, many of these approaches rely on fixed viewpoints (e.g., \textit{front, back, left, right}), limiting their applicability in real-world scenarios.

Another line of research explores pose-free 3D reconstruction using uncalibrated images as direct input. Several approaches~\cite{lin2024relpose++, jiang2024few} regress camera poses through network predictions,
while PF-LRM~\cite{wang2023pf} adapts LRM by incorporating a differentiable PnP module to predict poses from multi-view images for 3D reconstruction. iFusion~\cite{wu2023ifusion} leverages Zero123~\cite{liu2024one} predictions within an optimization-based pipeline to align poses. SpaRP~\cite{xu2024sparp} employs a coordinate-map representation with a generative diffusion model but relies on an additional PnP solver for refinement and is limited to no more than 6 views. In contrast, our regression-based method accommodates an arbitrary number of unposed inputs, providing a more efficient rendering pipeline while maintaining high-quality results for practical 3D reconstruction.

\section{Method}\label{sec:method}
LucidFusion is a feed-forward 3D reconstruction model that processes one to $N$ unposed images, recovering pose and object Gaussians. 
In Sec.~\ref{sec:preliminary}, we first examine how existing reconstruction models are formulated. Building on these insights, Sec.~\ref{sec:RCM} introduces the \textit{Relative Coordinate Map (RCM)}, a novel representation directly regressed from input images that enables pose estimation and 3D reconstruction without explicit pose information. Sec.\ref{sec:3dgs} extends RCM into \textit{Relative Coordinate Gaussians (RCG)} via 3D Gaussian Splatting~\cite{kerbl20233d}, enforcing global 3D consistency through a rendering loss. Finally, Sec.~\ref{sec:loss} presents our two-stage training strategy for efficient  3D reconstruction.

\begin{figure}[!t]
    \centering
    \includegraphics[width=\linewidth]{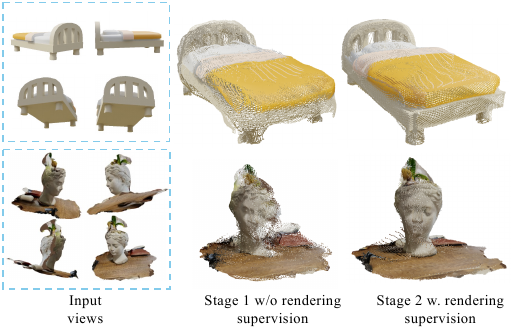}
    \captionsetup{font=small}
    \caption{Visualization of stage 1 and stage 2 results. }
    \label{fig:stage_1_2_pc_compare}
\end{figure}

\subsection{Preliminary}
\label{sec:preliminary}  

Extending a reconstruction pipeline from a single image to multiple images introduces several challenges. We abstract the 3D reconstruction problem as a mapping task: with a single image, the primary goal is to extract geometric information for object generation, whereas with multiple images, both mapping and scaling issues arise. This mapping can be performed explicitly, as in traditional MVS-based methods~\cite{yao2018mvsnet,chen2021mvsnerf}, or implicitly, as in LRM-based approaches~\cite{hong2023lrm}. However, both strategies typically rely on pose estimation, where images must either be pre-posed or restricted to specific viewpoints, limiting the pipeline's flexibility. In contrast, we propose a method that performs the mapping end-to-end without relying on explicit pose information.

We argue that a key challenge in multi-view reconstruction is ensuring consistent geometric feature estimation across different viewpoints, while also preserving scale-wrapping relationships. From this perspective, pose is merely an intermediate variable that performs the mapping. If pose information is embedded in the regression objective itself, it can be bypassed, thereby improving overall usability and reducing the pipeline’s complexity.

Building on this idea, \textit{Canonical Coordinate Map (CCM)}~\cite{li2023sweetdreamer} represents a natural approach by embedding pose information directly into an image’s pixel values. However, when regressing CCM from multi-view inputs, the model must operate under a world-coordinate convention and therefore simultaneously infer both orientation and geometry. This limitation becomes evident in our pilot study, where we regress a model using CCM (see the middle row of Fig.~\ref{fig:ccm_vs_rcm}): the same object parts—such as a sheep’s head and tail—should retain consistent colors across all views. This semantic information is crucial for indicating an object's orientation in world space. Any misalignment suggests that the model fails to accurately align the 2D multi-view inputs in 3D space.

\subsection{Relative Coordinate Map}
\label{sec:RCM}
For a reconstruction task, however, it is more important to maintain 3D consistency across input views than to learn an object’s canonical orientation. Hence, we propose the \textit{Relative Coordinate Map (RCM)}, which transforms each view’s coordinates to align with the coordinate system of a selected “main” view. As shown in the bottom row of Fig.~\ref{fig:ccm_vs_rcm}, this transformation resolves orientation ambiguities in our pilot study, making it more suitable for the reconstruction task.

Let \( \{\bm{I}_{i}\}^{N}_{i=1} \) be a set of $N$ input images, each \( \bm{I}_{i} \in \mathbb{R}^{H \times W \times 3} \). We define \textit{RCM} for each image as \(\bm{M}_{i} \in \mathbb{R}^{H \times W \times 3} \), where $\bm{M}_{i}$ contains the 3D coordinates corresponding to each pixel in $\bm{I}_{i}$. To help the model learn these coordinates from arbitrary viewpoints, we project all $N$ images into the coordinate system of a randomly chosen input view. This random selection encourages generalization of different viewpoints.

Concretely, for each input view, we have a camera pose \(\bm{P}_{i} \in \mathbb{R}^{4 \times 4}\) and an intrinsic matrix \(\bm{K} \in \mathbb{R}^{4 \times 4}\) (both in homogeneous form), as well as a depth map \(\bm{D}_{i} \in \mathbb{R}^{H \times W}\). We then randomly select one of these poses, \(\bm{P}_{main}\), as the main camera pose. We define the main camera's RCM as: 
\begin{equation}
    \bm{M}_{main} = \bm{P}_{main}\bm{P}^{-1}_{main}\bm{K}^{-1} \ast \bm{D}_{main},
\end{equation}
which simplifies to 
\begin{equation}
    \bm{M}_{main} = \bm{K}^{-1} \ast \bm{D}_{main},
    \label{eq:RCM_main}
\end{equation}
within its own camera coordinate frame. For remaining $N-1$ views, we transform each one into the main camera's coordinates:
\begin{equation}
    \bm{M}_{j} = \bm{P}_{main}\bm{P}^{-1}_{j}\bm{K}^{-1} \ast \bm{D}_{j}, \quad j = 1, 2, 3, \dots, N-1,
    \label{eq:RCM}
\end{equation}
with the RCM values constrained to $[-1, 1]$. To further enforce 3D consistency across multiple views, we concatenate all input images along the width dimension $W$, allowing the model to use self-attention to integrate multi-view information.

The RCM representation offers several key advantages. First, as an \textit{image-based} representation, it benefits from pre-trained foundation models, thereby simplifying the learning process. Second, RCM preserves a one-to-one mapping between image pixels and their corresponding 3D points, effectively capturing the geometry as a point cloud. Finally, since each RCM explicitly represents the position (x,y,z) of every pixel, we can compute the pose $\xi_{i}$ for each view $M_{i}$ using a standard Perspecitive-n-Point (PnP) solver~\cite{opencv_library}, enabling relative pose estimation. 

\subsection{Relative Coordinate Gaussians}\label{sec:3dgs}

Building on relative coordinate maps, one could train a 2D image-to-image model directly for unconstrained 3D reconstruction. However, we observe that naively performing this mapping often results in inconsistent and noisy outputs as shown in Fig.~\ref{fig:stage_1_2_pc_compare}, primarily due to the lack of 3D prior supervision that is crucial for maintaining 3D consistency. To address this, we integrate 3D Gaussians~\cite{szymanowicz2024splatter} with the relative coordinate map, forming what we call the \textit{Relative Coordinate Gaussians (RCG)}.

Specifically, we take the relative coordinates as the center of each Gaussian point. Beyond simply regressing the 3D position, we also regress the Gaussian parameters. Since the RCG is pixel-aligned, we can seamlessly expand the network’s output channels from 3 to 14. These additional channels encode the scale $\bm{s}$ (3 channels), the rotation quaternion $\bm{rot}$ (4 channels), and the opacity $\bm{\sigma}$ (1 channel). With these Gaussian parameters, we employ differentiable rasterization from arbitrary viewpoints, supervised by ground-truth images rather than solely by per-view coordinate predictions. This global rendering loss enforces consistency across views and yields smoother, more coherent reconstructions, as shown in Fig.~\ref{fig:stage_1_2_pc_compare}.

\subsection{Two Stage Training}\label{sec:loss}
\begin{figure}[!t]
    \centering
    \includegraphics[width=\linewidth]{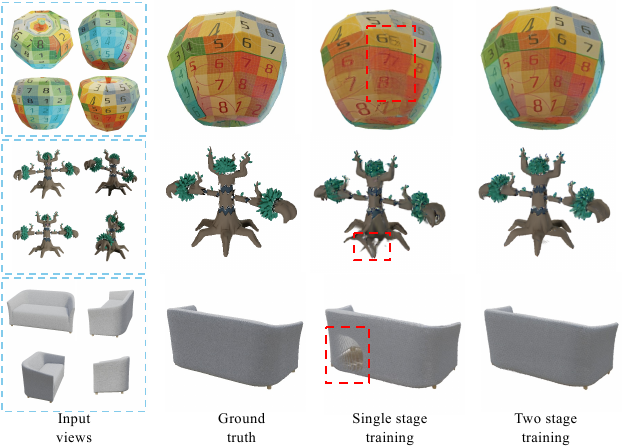}
    \captionsetup{font=small}
    \caption{Comparison with single and two-stage training. For single stage, the model struggles to predict Gaussian locations.}
    \label{fig:ablation1}
\end{figure}

We observe that jointly optimizing both the Relative Coordinate Map (RCM) and the rendering objective often leads to training instability. As illustrated in Fig.~\ref{fig:ablation1}, the network fails to localize the object geometry accurately and maintain multi-view consistency, resulting in misalignments or empty holes of the object. This arises because the model must simultaneously reason about per-pixel alignment and global 3D consistency, creating conflicting objectives during training.
To overcome this challenge, we adopt a two-stage training scheme. In Stage 1, we train the network on the RCM representation and using stable diffusion–based prior similar to~\cite{he2024lotus}, enabling it to learn robust mappings from the input images to the RCM. In Stage 2, we expand the learned RCM into the RCG representation and incorporate a differentiable rendering loss to enforce 3D consistency. By decoupling these learning stages, we alleviate the tension between local pixel alignment and global geometry constraints, substantially stabilizing the training process.

\paragraph{Stage 1.} We try to train a network to learn RCM represention. Let $E$ be the network mapping $N$ RGB images $\{\mathbf{I}_{i}\}^{N}_{i=1}$, where \(\mathbf{I}_{i} \in \mathbb{R}^{H \times W \times 3} \), to their corresponding RCMs \(\mathbf{M}_{i} \in \mathbb{R}^{H \times W \times 3}\). Formally,
\begin{equation}
    \mathbf{\hat{M}_{i}} = E(\mathbf{I}_{i}).
    \label{eq:encoder}
\end{equation}
We obtain ground truth RCMs from Eq.~\ref{eq:RCM} and supervise the predicted RCMs $\mathbf{\hat{M}_{i}}$ via MSE loss:
\begin{equation}
    \mathcal{L}_{rcm} = \frac{1}{N}\sum^{N}_{i=1}\mathcal{L}_{MSE}(\mathbf{\hat{M}}_{i}, \mathbf{M}_{i}).
    \label{loss:stage_1}
\end{equation}
After Stage 1, the network $E$ serves as a base model that reliably transforms input images into RCMs.

\paragraph{Stage 2.}
We then extend the output layer to introduce RCGs as Sec.~\ref{sec:3dgs}.
Specifically, We extract an intermediate feature map \(\mathbf{f}_{i} \in \mathbb{R}^{\frac{H}{8} \times \frac{W}{8} \times l}\) from $E$, which is passed to a decoder $G$ to predict the 14-channel RCGs $\mathbf{\Theta}_{i}$:
\begin{align}
    \mathbf{\Theta}_{i} &= G(\mathbf{f}_{i}), \label{eq:decoder} \\
    \mathbf{\Theta}_{i} &= (\mathbf{\hat{M}_{i}}, \mathbf{I}_{i} + \bm{\delta}^{c}_{i}, \mathbf{s}_{i}, \mathbf{rot}_{i}, \bm{\sigma}_{i}). \label{eq:gs_splats}
\end{align}

We render $I_{i}$ supervision views using a differentiable renderer~\cite{kerbl20233d}, and supervise it with its ground-truth view $\mathbf{I}_{i}$. To enforce visual fidelity, we adopt a combination of MSE loss, SSIM loss from~\cite{kerbl20233d}, and VGG-based LPIPS loss~\cite{zhang2018unreasonable}:
\begin{equation}
\begin{split}
    \mathcal{L}_{rgb} &= (1-\lambda)\mathcal{L}_{MSE}(\mathbf{\hat{I}}_{i}, \mathbf{I}_{i}) \\
    &\quad + \lambda \mathcal{L}_{SSIM}(\mathbf{\hat{I}}_{i}, \mathbf{I}_{i}) \\
    &\quad + \mathcal{L}_{LIPIS}(\mathbf{\hat{I}}_{i}, \mathbf{I}_{i}),
\end{split}
\label{eq:loss_rgb}
\end{equation}
where $\lambda = 0.2$, following~\cite{kerbl20233d}. To further accelerate convergence and enhance object boundaries, we also apply an MSE loss to the alpha channel~\cite{tang2024lgm}:
\begin{equation}
    \mathcal{L}_{\alpha} = \mathcal{L}_{MSE}(\mathbf{\hat{I}}^{\alpha}_{i}, \mathbf{I}^{\alpha}_{i}).
    \label{eq:loss_alpha}
\end{equation}
Thus, the overall loss for Stage 2 is given by
\begin{equation}
    \mathcal{L} = \mathcal{L}_{rgb} + \mathcal{L}_{\alpha}.
    \label{eq:loss_stage2}
\end{equation}

\begin{figure*}[!t]
    \centering
    \includegraphics[width=\textwidth]{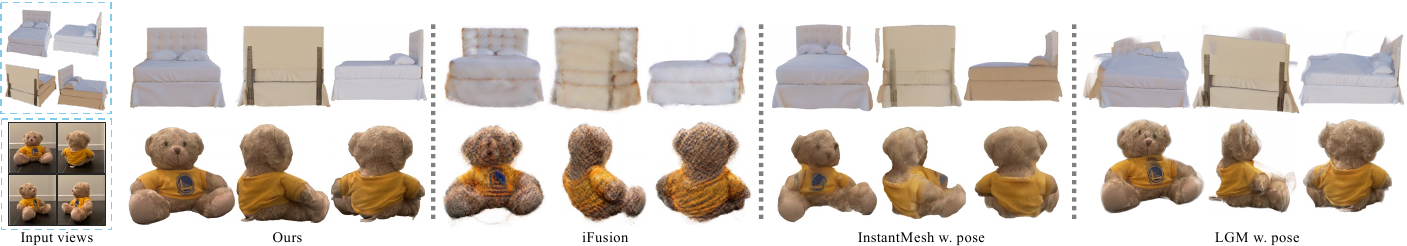}
    \captionsetup{font=small}
    \caption{Qualitative comparison with iFusion~\cite{wu2023ifusion}, InstantMesh~\cite{xu2024instantmesh} and LGM~\cite{tang2024lgm} under sparse view setting.}
    \label{fig:qualitative-random-view}
\end{figure*}

\begin{figure*}[!htpb]
    \centering
    \includegraphics[width=\textwidth]{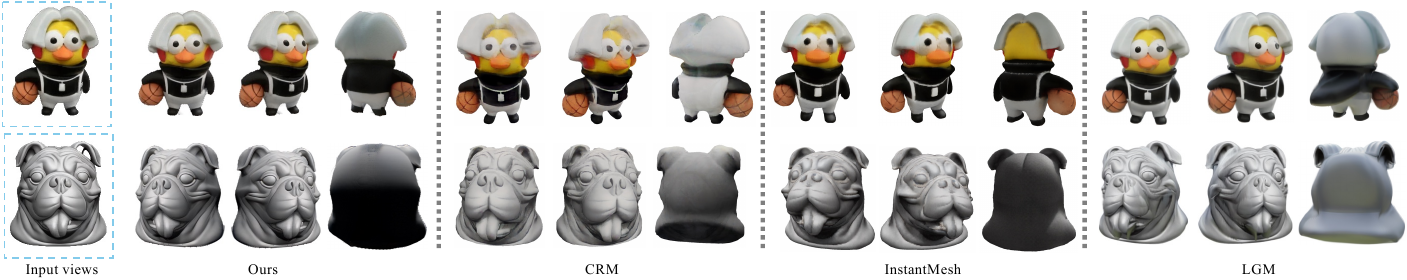}
    \captionsetup{font=small}
    \caption{Qualitative comparison with InstantMesh~\cite{xu2024instantmesh}, CRM~\cite{wang2024crm} and LGM~\cite{tang2024lgm} under standard single-image-to-3D paradim.}
    \label{fig:qualitative-single-view}
\end{figure*}

\paragraph{Pose Estimation.} As we disccussed, since the central of RCG are defined as the spacial coordinates of each pixel, we can estimate the camera pose by minimizing the reprojection error of 3D–2D point correspondences. Assume $\mathbf{q}_{i,j}$ represents 3D point location (x,y,z) in RCM view $i$, and $\mathbf{p}_{i,j}$ represents 2D pixel location at $j$ of the RCM view $i$, we have:  
\begin{equation}
    \xi_{i} = arg min\sum_{j=1}^{N}||Proj(R_{i}\cdot \mathbf{q}_{i,j} + t_{i}) - \mathbf{p}_{i,j}||^{2} ,
    \label{eq:pnp}
\end{equation}
where $R_{i}, t_{i}$ are the rotation and translation matrix, and $N$ represents number of pixels in each of the RCM $M_{i}$. We use RANSAC scheme in OpenCV~\cite{opencv_library} and filter out non-informative white background points from affecting the pose prediction. We present these results in Sec.~\ref{sec:exp}.

\begin{figure*}[!t]
    \centering
    \includegraphics[width=\textwidth]{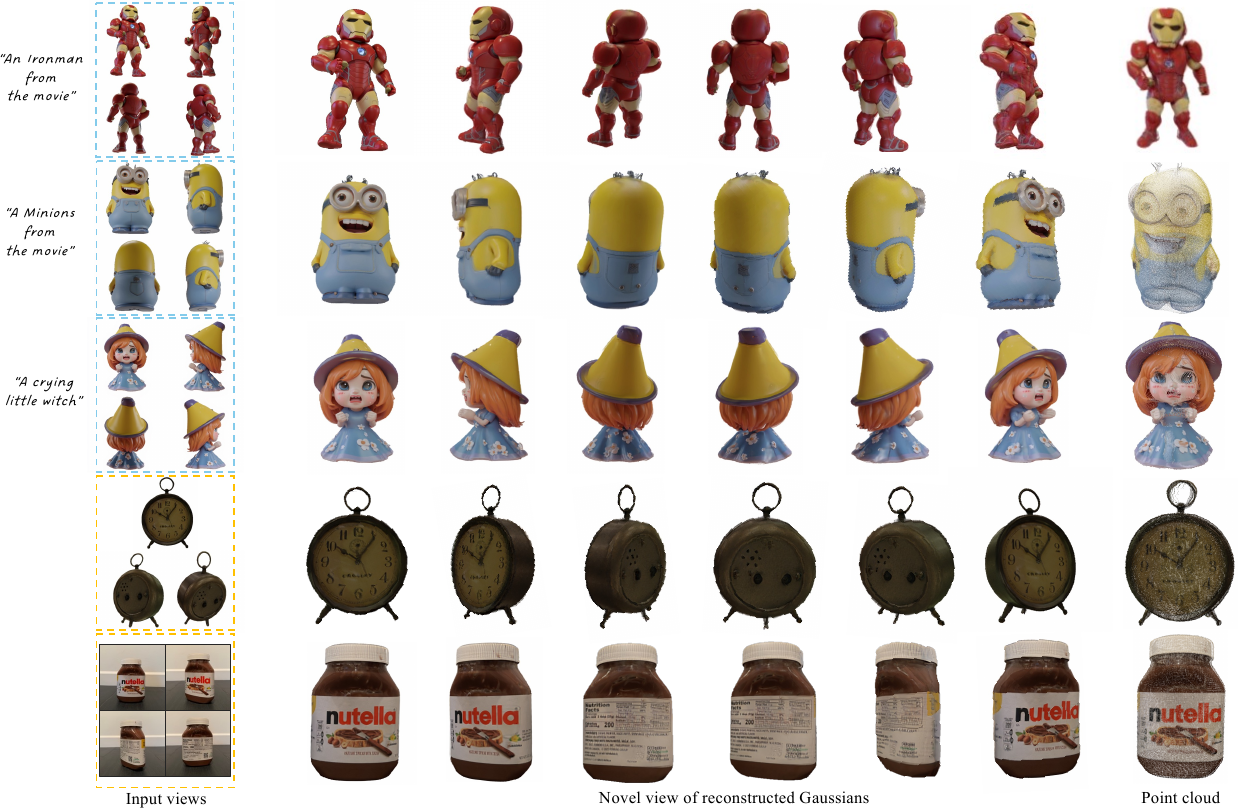}
    \captionsetup{font=small}
    \caption{Our method generates high-resolution 3D Gaussians on various input types. The blue box shows input views are generated using a Text-to-Image diffusion model, where the yellow box shows sparse view inputs.}
    \label{fig:qualitative-main}
\end{figure*}

\section{Experiment}
In this section, we first explain our training and testing datasets in Sec.~\ref{sec:exp_setting}. We then make both quantitative and qualitative comparisons against different baselines in Sec.~\ref{sec:exp}. Finally, we explain our design choice in Sec.~\ref{sec:ablation}.

\begin{table}[]
\centering
\resizebox{\linewidth}{!}{%
\begin{tabular}{@{}cccccc@{}}
\toprule
Dataset               & Method     & Rot. error$\downarrow$ & Acc. @$15^\circ\uparrow$ & Acc. @$30^\circ\uparrow$ & T.error$\downarrow$       \\ \midrule
\multirow{3}{*}{GSO}  & RelPose++~\cite{lin2024relpose++}  & 101.24         & 0.014         & 0.087         & 1.75          \\
                      & iFusion~\cite{wu2023ifusion}    & 107.29         & 0.011         & 0.086         & 1.05          \\
                      & Ours & \textbf{11.50} & \textbf{0.93} & \textbf{0.99} & \textbf{0.16} \\ \midrule
\multirow{3}{*}{ABO}  & RelPose++~\cite{lin2024relpose++}  & 103.23         & 0.016         & 0.092         & 1.74          \\
                      & iFusion~\cite{wu2023ifusion}    & 102.68         & 0.016         & 0.094         & 1.13          \\
                      & Ours & \textbf{19.40} & \textbf{0.77} & \textbf{0.84} & \textbf{0.17} \\ \midrule
\multirow{3}{*}{OO3D} & RelPose++~\cite{lin2024relpose++}  & 104.23         & 0.017         & 0.092         & 1.78          \\
                      & iFusion~\cite{wu2023ifusion}    & 106.95         & 0.012         & 0.086         & 1.18          \\
                      & Ours & \textbf{12.91} & 0.85          & \textbf{0.97} & \textbf{0.13} \\ \bottomrule
\end{tabular}%
}
\caption{Performance on pose prediction task. We compare cross-dataset generalization on GSO~\cite{downs2022google}, ABO~\cite{collins2022abo} and OminiObject3D (OO3D)~\cite{wu2023omniobject3d} with baselines RelPose++~\cite{lin2024relpose++}, iFusion~\cite{wu2023ifusion}.}
\label{tab:pose_predict}
\end{table}

\subsection{Experimental Setting}\label{sec:exp_setting}
We train our model on a subset of Objaverse~\cite{deitke2023objaverse} dataset as there are many low quality 3D shapes in the original set. The final training data contains approximately 98K 3D objects. For each 3D object, we generate a total of 90 views with different elevations. During training, $N$ views are randomly sampled from these 90 images. The rendered images have a resolution of $512\times 512$ and are generated under uniform lighting conditions.

To evaluate our model's generalization ability cross different datasets, we utilize GSO~\cite{downs2022google}, ABO~\cite{collins2022abo} and OmniObject3D~\cite{wu2023omniobject3d}. We randomly choose 200 objects for evaluating our model's performance given sparse images as input. For each object, we randomly render 24 views at different elevations, and randomly-chosen 4 of them as our input images to our model to predict pose and novel view rendering quality. More details please see the Appendix.

\begin{table*}[htbp]
    \centering
    \begin{minipage}[t]{0.48\textwidth} 
        \centering
        \small
        \resizebox{\textwidth}{!}{%
        \begin{tabular}{@{}lcccccc@{}}
        \toprule
                    & \multicolumn{3}{c|}{GSO}        & \multicolumn{3}{c}{ABO}        \\ 
                    & PSNR$\uparrow$ & SSIM$\uparrow$  & \multicolumn{1}{c|}{LPIPS$\downarrow$} & PSNR$\uparrow$ & SSIM$\uparrow$  & LPIPS$\downarrow$ \\ \midrule
        iFusion~\cite{wu2023ifusion}         & 17.21              & 0.852       & \multicolumn{1}{c|}{0.180}     & 17.54              & 0.853     & 0.180     \\
        LGM~\cite{tang2024lgm}     & 19.61          & 0.872           & \multicolumn{1}{c|}{0.131}     & 19.89              & 0.873     & 0.131 \\
        InstantMesh~\cite{xu2024instantmesh} & 20.75              & 0.894   & \multicolumn{1}{c|}{0.127}     & 20.98              & 0.901     & 0.129  \\
        Ours            & \textbf{25.97} & \textbf{0.930}  & \multicolumn{1}{c|}{\textbf{0.070}} & \textbf{25.98} & \textbf{0.917} & \textbf{0.088} \\
        \bottomrule
        \end{tabular}%
        }
        \captionsetup{font=small}
        \caption{Performance comparison against baselines on GSO~\cite{downs2022google} and ABO~\cite{collins2022abo} for 4 views input.}
        \label{tab:table_1}
    \end{minipage}
    \hfill 
    \begin{minipage}[t]{0.48\textwidth} 
        \centering
        \small
        \resizebox{\textwidth}{!}{%
        \begin{tabular}{@{}lcccccc@{}}
        \toprule
                    & \multicolumn{3}{c|}{GSO}        & \multicolumn{3}{c}{ABO}        \\ 
                    & PSNR$\uparrow$ & SSIM$\uparrow$  & \multicolumn{1}{c|}{LPIPS$\downarrow$} & PSNR$\uparrow$ & SSIM$\uparrow$  & LPIPS$\downarrow$ \\ \midrule
        CRM~\cite{wang2024crm}             & 16.74          & 0.858               & \multicolumn{1}{c|}{0.177}     & 19.23              & 0.871     & 0.169     \\
        LGM~\cite{tang2024lgm}             & 14.31          & 0.824           & \multicolumn{1}{c|}{0.186} & 16.03          & 0.861 & 0.181 \\
        InstantMesh~\cite{xu2024instantmesh}     & 16.84           & \textbf{0.864}               & \multicolumn{1}{c|}{0.177}     & \textbf{19.73}              & 0.873     & 0.168 \\
        Ours            & \textbf{16.91} & 0.862  & \multicolumn{1}{c|}{\textbf{0.177}} & 19.51 & \textbf{0.873} & \textbf{0.168} \\
        \bottomrule
        \end{tabular}%
        }
        \captionsetup{font=small}
        \caption{Performance comparison against baselines on GSO~\cite{downs2022google} and ABO~\cite{collins2022abo} for single-image-to-3D setting.}
        \label{tab:table_single_to_3d}
    \end{minipage}
    \vspace{-0.1in}
\end{table*}

\subsection{Experiment Results}\label{sec:exp}
\paragraph{Reconstruction.} We first compare our methods under sparse view settings with a recent open-source pose-free method iFusion~\cite{wu2023ifusion}, and feed-forward based methods LGM~\cite{tang2024lgm} and InstantMesh~\cite{xu2024instantmesh}. Since LGM and InstantMesh can only work when camera poses are given, we supply them with the ground truth pose for 3D reconstruction. We report PSNR, SSIM and LPIPS metrics for measuring the image quality. As we show in Tab.~\ref{tab:table_1}, our model consistently outperforms baselines with a large margin. In addition, we visualize the result in Fig.~\ref{fig:qualitative-random-view}, where the top row shows an object from ABO~\cite{collins2022abo}, and the bottom row shows an in-the-wild captured image. For the in the wild capture, we supply LGM and InstantMesh with our predicted poses, we notice that even with poses, LGM and InstantMesh still struggle to reconstruct the object, as they have overfitted their inputs to the fixed camera position, where our method gives better geometry and visual quality.

We then follow the standard single-image-to-3D paradigm to evaluate our method to demonstrate the flexibility of our method with the standard approach. Specifically, we use the off-shelf Flux~\cite{flux2024} diffusion model to generate multi-views. As shown in Tab.~\ref{tab:table_single_to_3d}, our approach works effectively within the existing single-image-to-3D paradigm, delivering on-par performance with current baselines. In addition, as we show in Fig.~\ref{fig:qualitative-single-view}, our method can utilize the multi-view diffusion model and faithfully produce results at $512\times512$. 

In Fig.~\ref{fig:qualitative-main}, we demonstrate our model's generalization ability across different data sources. In the top 3 rows, we showcase where our method pairs with a Text-to-Image (T2I) Flux~\cite{flux2024} model, and the bottom two rows we show results with scanned object and in-the-wild captured object. Our model produces high-quality results at a resolution of $512 \times 512$, demonstrates the capability for real-world applications with only arbitrary number of sparse input views.

\paragraph{Pose Estimation.} We compare LucidFusion with feed-forward approach RelPose++~\cite{lin2024relpose++} and recently open sourced optimization based approach iFusion~\cite{wu2023ifusion}. We follow iFusion~\cite{wu2023ifusion} and measure median error in rotation and translation. We also report the relative rotation accuracy below thresholds $15^\circ$ and $30^\circ$. 
As shown in Tab.~\ref{tab:pose_predict}, our method consistently outperforms baseline models across different datasets. It is worth noting that optimization-based pipeline iFusion introduces a 5 mins optimization time for each of the objects, where our method recovers pose and object shapes with a single feed-forward pass, demonstrating our superior performance for real-world applications. Please see more results in Appendix.

\subsection{Ablation Study}\label{sec:ablation}

\begin{figure}[!t]
    \centering
    \includegraphics[width=\linewidth]{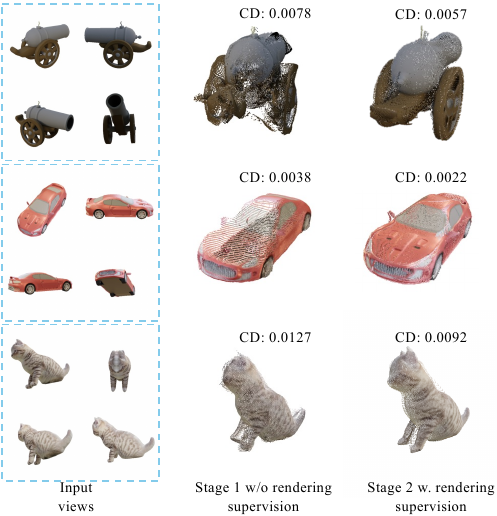}
    \captionsetup{font=small}
    \caption{Point cloud visualization for stage 1 and 2. We also show their chamfer distance.}
    \label{fig:ablation chamfer distance}
    \vspace{-0.2in}
\end{figure}

\paragraph{Importance of RCG.} As detailed in Sec.~\ref{sec:3dgs}, the RCG representation enforces global 3D consistency across all input views. Fig.~\ref{fig:ablation chamfer distance} illustrates several examples of point clouds derived from the RCG representation, where we extract both position and RGB data for visualization. We also compute the Chamfer Distance between each stage’s output and the ground-truth point cloud. The results clearly show that incorporating the RCG representation produces smoother, more coherent reconstructions.

\paragraph{Training Scheme.} As we explained in Sec.~\ref{sec:loss}, jointly optimizing the model with RCM and rendering supervision leads to misalignment and empty holes, as the network struggles to localize the object geometry and maintain multi-view consistency, as we show in Fig.~\ref{fig:ablation1}. These artifacts reflect the incorrect position extracted from the predicted RCGs. However, in the two-stage training scheme, we first learn the per-pixel alignment using the RCM supervision, and extend the RCM to RCG to utilize rendering supervision to ensure 3D consistency across multi-views.

\begin{figure}[!t]
    \centering
    \includegraphics[width=\linewidth]{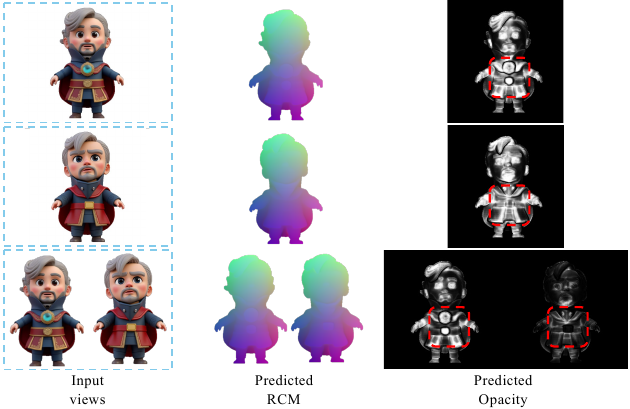}
    \captionsetup{font=small}
    \caption{Visualization of predicted RCM and opacity extracted from RCG as confidence map.}
    \label{fig:conf_conflict}
    \vspace{-0.2in}
\end{figure}

\paragraph{Gaussians Opacity as Confidence.} As we illustrated in Sec.~\ref{sec:3dgs}, extending RCM to RCG not only enables supervision via rendering but also enforces 3D consistency across views, leading to a globally optimized 3D representation. Without RCG refinement, multi-view misalignment can cause conflicts, as the model maps image pixels directly to 3D points, leading to geometric ambiguities. However, RCG’s opacity serves as a confidence measure, filtering out conflicting regions across input images and improving multi-view fusion.
As shown in Fig.~\ref{fig:conf_conflict}, tthe predicted opacity maps reflect confidence in different regions, allowing the model to lower opacity in conflicting areas and preserve object rendering quality.

\subsection{Limitation}
Despite the promising results, our model has some limitations. First, it can only render objects positioned at the center of the scene, without backgrounds. We hypothesize that incorporating background information into the RCG representation during training could address this issue, which we leave for future work. Additionally, our current model is trained only with Objaverse data. Future work could explore training on a wider variety of settings to enhance the robustness of the RCG representation.

\section{Conclusion} 
In this work, we propose LucidFusion, a flexible end-to-end feed-forward framework that leverages the \textit{Relative Coordinate Gaussians (RCG)}, a novel representation designed to align geometric features coherently across different views. Our model first maps RGB inputs to \textit{Relative Coordinate Map (RCM)} representations and extended it to RCG for simultaneously reconstructing the object and recover pose, all in a feedforward manner.  This approach alleviates the pose requirements in the 3D reconstruction pipelines, and delivers high-quality outputs across a range of scenarios. LucidFusion can also integrate seamlessly with the original single-image-to-3D pipeline, making it a versatile tool for 3D object reconstruction. We hope this work will open new avenues for future research in the field of 3D reconstruction.  
{
    \small
    \bibliographystyle{ieeenat_fullname}
    \bibliography{main}
}

\end{document}